\documentclass[runningheads]{llncs}
\usepackage{graphicx}
\usepackage{subfiles}
\usepackage{amsmath,amssymb} 
\usepackage{color}
\usepackage[width=122mm,left=12mm,paperwidth=146mm,height=193mm,top=12mm,paperheight=217mm]{geometry}
\begin{document}
\title{Learning image-to-image translation using paired and unpaired training samples}
\titlerunning{Paired and unpaired image-to-image translation}
%
\author{Soumya Tripathy \inst{1}\and
Juho Kannala\inst{2}\and
Esa Rahtu\inst{1} }
\authorrunning{Tripathy et al.}
%
\institute{Tampere University of Technology\\
 Tampere, Finland \\
\email{\{soumya.tripathy,esa.rahtu\}@tut.fi}\and
Aalto University of Technology\\
  Helsinki, Finland \\
\email{juho.kannala@aalto.fi}}
\maketitle              
\begin{abstract}
Image-to-image translation is a general name for a task where an image from one domain is converted to a corresponding image in another domain, given sufficient training data. Traditionally different approaches have been proposed depending on whether aligned image pairs or two sets of (unaligned) examples from both domains are available for training. While paired training samples might be difficult to obtain, the unpaired setup leads to a highly under-constrained problem and inferior results. In this paper, we propose a new general purpose image-to-image translation model that is able to utilize both paired and unpaired training data simultaneously. We compare our method with two strong baselines and obtain both qualitatively and quantitatively improved results. Our model outperforms the baselines also in the case of purely paired and unpaired training data. To our knowledge, this is the first work to consider such hybrid setup in image-to-image translation.    
\end{abstract}

\section{Introduction}
\label{sec:intro}
Many vision and graphics problems such as converting grayscale to colour images \cite{c37}, satellite images to maps \cite{c3}, sketches to photographs \cite{c36}, and paintings to photographs \cite{c1} can be seen as image-to-image translation tasks. Traditionally, each problem has been tackled with specific architectures and loss functions that do not generalize well to other tasks. 
Recently, Generative Adversarial Networks (GANs) \cite{c10} have provided an alternative approach, where the loss function is learned from the data instead of defining it manually. Impressive results have been demonstrated in several areas such as image inpainting \cite{c38}, style transfer \cite{c1}, super-resolution \cite{c13} and image colorisation \cite{c6}. Still, only a few papers consider a general purpose solution. One such work is \cite{c3} that proposes a conditional GAN architecture applicable to a wide range of translations tasks. Unfortunately, the system requires a large set of input-output image pairs for training. While such data might be easy to obtain for certain tasks (e.g. image colorization), it poses a major restriction in general. 

\begin{figure}[t]
\centering
\includegraphics[width=12.2cm,height=5.3cm]{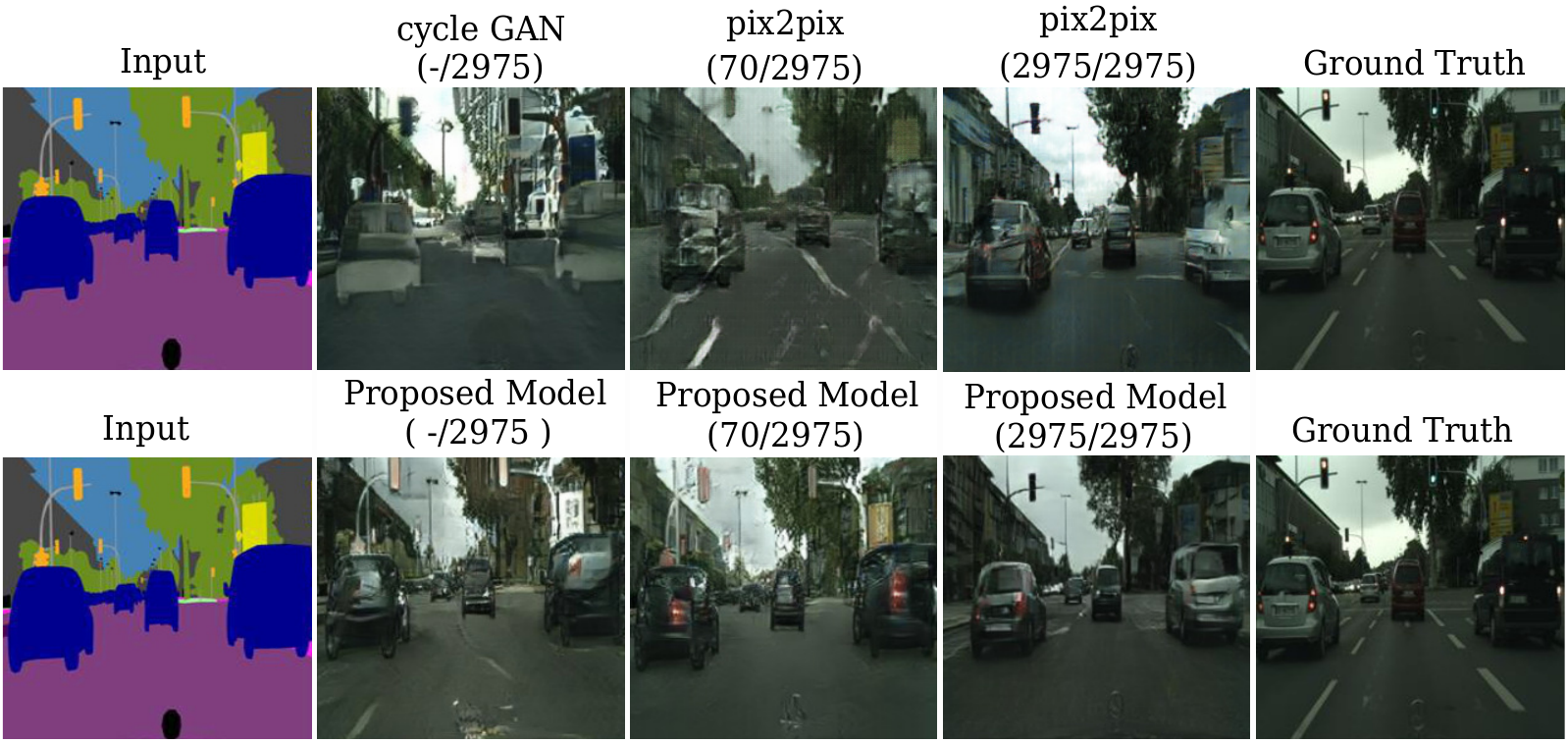}
\caption{An example result for label map to photograph translation task. From left to right (columnwise): input label map, cycleGAN \cite{c1} vs proposed model (only unpaired training examples), pix2pix \cite{c3} vs proposed model (70 paird data out of 2975 training samples), pix2pix \cite{c3} vs proposed model (2975 paired training samples) and the ground truth. Our hybrid model provides a clear benefit over the baseline methods.}
\label{fig1}
\end{figure}

To overcome this limitation, several works \cite{c1,c2,c20,c21,c22,c24,c25} have proposed models that learn the mapping using only a set of (unpaired) examples from both domains (e.g.~a set of photographs and another set of paintings). Although suitable data is substantially easier to obtain, the problem becomes highly under-constrained. This clearly hampers the performance compared to models using paired training samples. For example, Figure \ref{fig1} shows semantic label to photograph translation results for \cite{c1}. It is quite evident that \cite{c1} gets confused between the labels as it produces buildings in the place of trees. This problem of label switching is very common in such models and it arises due to the lack of supervision. 

In practice, it would be beneficial to study a hybrid solution that is able to take advantage of both types training data. In many real-world applications, one can afford to produce a small set of paired training samples (e.g.~30) providing strong constraints for the task. The additional unpaired samples facilitate the learning of further details and improve the quality of the outputs. As shown in the Figure \ref{fig1}, our hybrid approach produces better quality results than the models using either paired or unpaired type of data during the training. Moreover, since there exists already a few datasets with paired training examples (e.g. semantic segmentation \cite{c5}, image colorization \cite{c6}, super-resolution \cite{c7}, and time hallucination \cite{c8}) it would make sense to leverage this information to slightly different but related problems. 

State of the art methods such as \cite{c1,c2} and \cite{c3,c19} are carefully designed to deal with unpaired and paired data respectively. For example, \cite{c3} utilizes conditional GAN framework \cite{c9} to take advantage of the paired training samples and \cite{c1} uses a cyclic structure with unconditional GANs to handle the unpaired examples during training. In practice, we found it difficult to use the existing models to take advantage of both types of training examples simultaneously. 

In this paper, we propose a new general purpose image-to-image translation approach that is able to utilize both paired and unpaired training examples. To our knowledge, this is the first work to study such hybrid problem. The additional contributions of our paper are 1) a new adversarial cycle consistency loss that improves the translation results also in purely unpaired setup, 2) a combination of structural and adversarial loss improving learning with paired training samples and 3) we demonstrate that it is possible to improve translation results by using training samples from related external dataset. 

\section{Related work}
\paragraph{Generative adversarial network (GAN)} 

The original GAN \cite{c10} consists of two competing network architectures called generator and discriminator. The generator network produces output samples which should ideally be indistinguishable from the training distribution. Since it is infeasible to engineer a proper loss function, the assessment is done by the discriminator network. Both networks are differentiable, which allows to train them jointly. Originally, GANs were mainly used to generate images from random inputs \cite{c11},\cite{c12},\cite{c13}. However, recent works also consider cases where generation is guided by an additional input such as semantic label map \cite{c9}, text \cite{c14,c15}, or attributes \cite{c16}. These kinds of models are generally known as conditional GANs \cite{c9}. 

\paragraph{Image-to-image translation using paired training data} 

Conditional GAN architectures have been widely applied in image-to-image translation tasks where sufficient amount of paired training examples are available \cite{c3,c17,c18,c19}. In particular, \cite{c3} proposed a general purpose image to image translator "pix2pix" for various applications such as semantic maps to real image, edges to shoe images, and gray to color images. This work was later extended in \cite{c19} to produce high-resolution outputs. In addition, some recent work have obtained excellent results without adversarial learning \cite{c26}. However, it would be difficult to generalize them to unpaired training samples.
\paragraph{Image-to-image translation using unpaired training data}  

Three concurrent works "cycleGAN" \cite{c1}, "DiscoGAN" \cite{c2} and "DualGAN" \cite{c20} propose architectures consisting of two generator-discriminator pairs. The first pair learns a forward mapping from source to target domain, while the second pair learns the inverse of it. Both pairs are trained simultaneously using adversarial losses and an additional loss that encourages cascaded mapping to reproduce the original input image (referred as cycle consistency loss in \cite{c1}). 

Recently, "CoGAN" \cite{c21} proposed a weight sharing strategy to learn a common representation across domains. This work was later extended in \cite{c22} by combining GANs and variational autoencoders \cite{c23}. "XGAN" \cite{c24} proposed a semantic consistency loss to preserve the semantic content of the image during the transfer. Finally, unpaired image-to-image translation is also studied without using adversarial loss \cite{c25}. Despite the recent progress, the results are still clearly inferior compared to methods using paired data (e.g. see Fig. \ref{fig1}).

\begin{figure}[t]
\centering
\includegraphics[width=11cm,height=6.5cm]{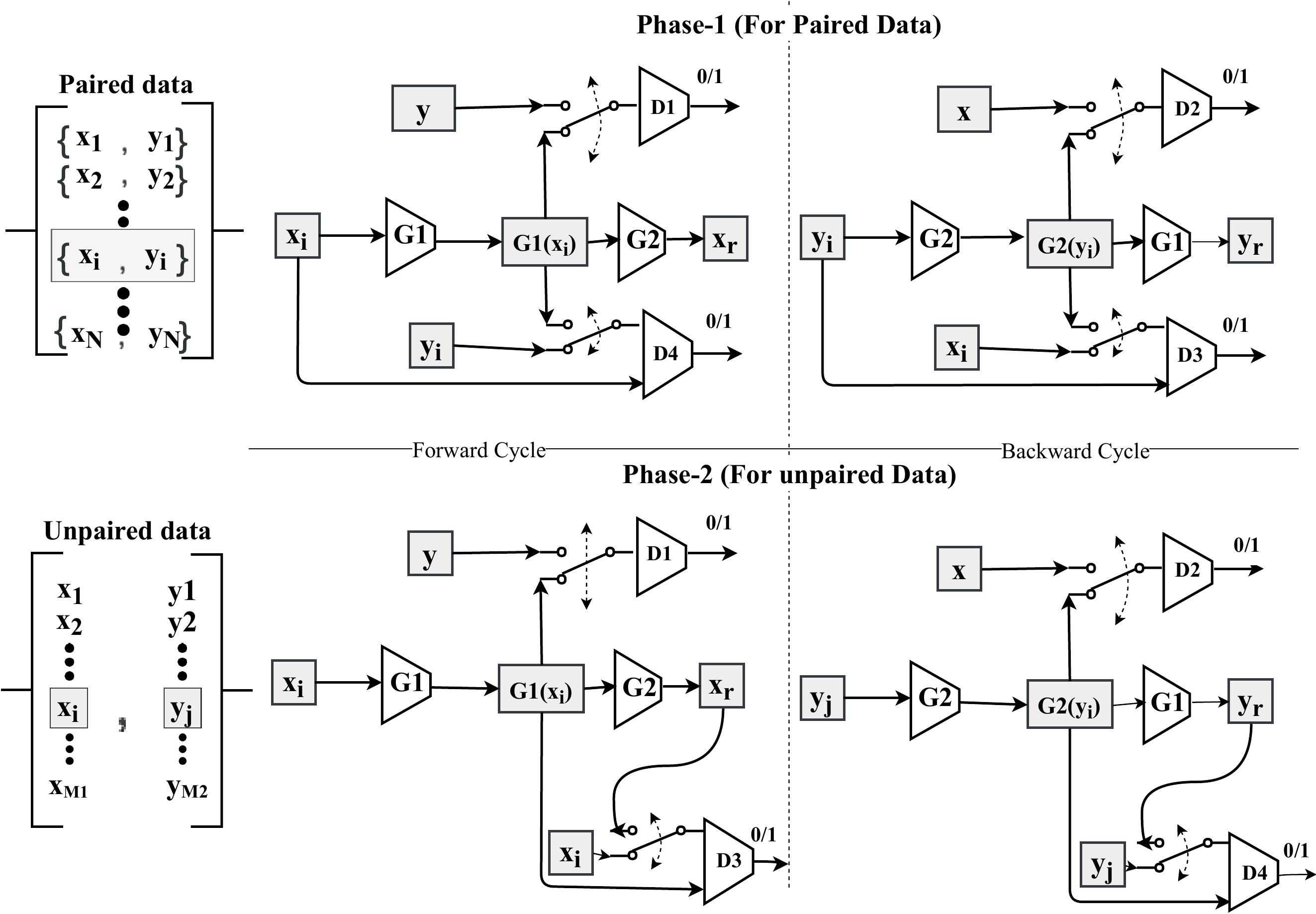}
\caption{The overall architecture of the proposed model for paired and unpaired training inputs. The two generators provide the forward and backward mappings \(G1 :X\rightarrow Y\) and \(G2 :Y\rightarrow X\). The discriminators $D1$ and $D2$ implement an adversarial loss encouraging the generators to produce outputs indistinguishable from the corresponding training distributions. The conditional discriminators $D3$ and $D4$ work in dual role by encouraging the translated images to be similar to ground truth for paired training samples and implementing an adversarial cycle consistency loss for unpaired training samples. }
\label{fig3}
\end{figure} 
\section{Method}
\vspace{-0.1cm}
\subsection{Task definition}

The overall task is to obtain a mapping from a source domain $X$ to a target domain $Y$. Figure \ref{fig1} illustrates an example case where $X$ and $Y$ correspond to semantic label maps and streetview photographs, respectively. The mapping is learned using a given training data, which in our case may consist of both aligned image pairs $\lbrace x_i,y_i  \rbrace_{i=1}^N$ and (unaligned) image samples $\lbrace x_i  \rbrace_{i=1}^{M_X}$ and $\lbrace y_j  \rbrace_{j=1}^{M_Y}$. Here $x_i \in X$, $y_j \in Y$, $N$ is the number of aligned samples, and $M_X$, $M_Y$ are the number of unpaired samples from each domain. Two interesting special cases occur if $N=0$ (fully unpaired setup) or $M_X=M_Y=0$ (fully paired setup). 
\subsection{Basic architecture}

The overall architecture of our solution is illustrated in Figure \ref{fig3}. The system contains two generator networks $G1$ and $G2$. The generator $G1$ provides a mapping from domain $X$ to domain $Y$ denoted as $G1 :X \rightarrow Y$ (e.g. semantic map to photograph), whereas the second generator learns the inverse of this mapping denoted as $G2 :Y \rightarrow X$ (e.g. photograph to semantic map). Generator networks are trained using adversarial loss \cite{c10} that is implemented by two discriminator networks $D1$ and $D2$. $D1$ aims to distinguish between transformation results $G1(x)$ and given training samples $y \in Y$. $D2$ works similarly for $G2(y)$ and $x \in X$. These loss functions will guide the generators to produce samples that are indistinguishable from the training image distributions $x \sim p_{data}(x)$ and $y \sim p_{data}(y)$ in corresponding domains. Following \cite{c10}, the exact formulation with respect to $G1$ and $D1$ is 
\scriptsize 
\begin{align}
\mathcal{L}_{GAN}(G1(X),D1(Y)) & =\mathbb{E}_{y}\left[ log D1(y)\right] + \mathbb{E}_{x}\left[ log( 1-D1(G1(x_i)))\right]
\label{D0n}
\end{align}
\normalsize 
where $x$ and $y$ are from data distribution $p_{data}(x)$ and $p_{data}(y)$ respectively. The corresponding objective $\mathcal{L}_{GAN}(G2(Y),D2(X)) $ for $G2$ and $D2$ is obtained similarly. Note that Equation (\ref{D0n}) does not depend on the pairing of the training samples. Therefore, this part of the loss function can be applied to all training samples in the same way.

\subsection{Cycle consistency}

Given only the discriminators $D1$ and $D2$, the learning problem remains poorly constrained and would not converge to a meaningful solution. Following \cite{c1}, we circumvent the problem by adding a cycle consistency loss that encourages the cascaded mapping to reproduce the original input. Mathematically, $G2(G1(x))  \approx x$ and $G1(G2(y)) \approx y$, where $x \in X$ and $y \in Y$. The intuition behind the cycle consistency is that if we translate from one domain to another and then back again we should arrive where we started (e.g. translating a sentence from one language to another and then back again). 

In \cite{c1}, the cycle consistency loss was defined as $L1$ difference between the reconstruction $G2(G1(x))$ and the input $x$ (similarly for $y$). Such loss would only be able to measure pixel level agreement between the images, although we might actually be more interested in semantic similarity (e.g. if you translate a sentence to another language and then back, the result might have the exact same meaning but expressed with different words). To this end, we propose a new adversarial cycle consistency loss implemented by two conditional discriminator networks $D3$ and $D4$ \cite{c9}. The dual input conditional discriminator $D3$ aims to determine if $x_r$ exactly matches $x_i$ given $G1(x_i)$. More formally:
\scriptsize 
\begin{eqnarray}
& \mathcal{L}_{cGAN}(G2(G1(X)),D3(G1(X),X)) = \mathbb{E}_{x}\left[ log D3(G1(x_i),x_i)\right]+ \mathbb{E}_{x}\left[ log( 1-D3(G1(x_i),x_r))\right]
\label{D3n}
\end{eqnarray}
\normalsize
This formulation encourages $x_i$ and $x_r$ to be similar in terms of image content. Similar expression can be written for $\mathcal{L}_{cGAN}(G1(G2(Y)),D4(G2(Y),Y))$ in backward cycle. 

\subsection{Conditional adversarial loss for aligned training samples}

In principle, the formulation (\ref{D3n}) would be applicable to both unpaired and paired training data. However, for paired training samples $(x_i,y_i)$ we can actually obtain much tighter constraint by enforcing the similarly between transformation result $G1(x_i)$ and the corresponding ground truth $y_i$. Now recall that the objective of the conditional discriminator $D4$ was to distinguish between real and reconstructed samples in domain $Y$. Therefore, the same network can be reused to discriminate between real ($y_i$) and generated image ($G1(x_i)$) in domain $Y$ given the input $x_i$. 
The corresponding loss function would be
\scriptsize 
\begin{align}
\mathcal{L}_{cGAN}(G1(X),D4(X,Y)) & =\mathbb{E}_{x,y}\left[ log D4(x_i,y_i)\right] + \mathbb{E}_{x}\left[ log( 1-D4(x_i,G1(x_i)))\right]
\label{D1n}
\end{align}
\normalsize
By utilising $D3$ and $D4$ in a dual role, we can obtain more compact architecture and efficient learning. Figure \ref{fig3} illustrates how our architecture works in the case of paired and unpaired image samples. 

\subsection{Complete learning objective}

The complete learning objective would be a composition of the presented loss functions. In order to further stabilise the learning process, we add weighted versions of the L1 cycle consistency loss and an identity loss \cite{c1} to our full objective. These are defined as follows:
\scriptsize 
\begin{align}
\mathcal{L}_{cyc \ell_1}(G1,G2) & =\mathbb{E}_{x}\left[ \parallel x_r-x_i\parallel_1 \right] + \mathbb{E}_{y}\left[ \parallel y_r-y_i\parallel_1 \right] \\
\mathcal{L}_{idt}(G1,G2) & =\mathbb{E}_{y}\left[ \parallel G1(y_i)-y_i\parallel_1 \right] + \mathbb{E}_{x}\left[ \parallel G2(x_i)-x_i\parallel_1 \right]
\label{D2n}
\end{align}
\normalsize
The complete loss function for paired training samples $\lbrace x_i,y_i  \rbrace_{i=1}^N$ and for the unpaired training samples $\lbrace x_i  \rbrace_{i=1}^{M_X}$ and $\lbrace y_j  \rbrace_{j=1}^{M_Y}$ are given in \eqref{fullobj1} and \eqref{fullobj2} respectively.
\scriptsize 
\begin{align}
\label{fullobj1}
 \mathcal{L}(G1,G2,D1,D2,D3,D4) = \mathcal{L}_{GAN}(G1(X),D1(Y))
+ \mathcal{L}_{GAN}(G2(Y),D2(X)) + \\
 \mathcal{L}_{cGAN}(G1(X),D4(X,Y))+ \mathcal{L}_{cGAN}(G2(Y),D3(X,Y)) + \lambda_{1}\mathcal{L}_{cyc \ell_1}(G1,G2) +  \lambda_2 \mathcal{L}_{idt}(G1,G2)\nonumber
\end{align}
\scriptsize 
\begin{align}
\label{fullobj2}
 \mathcal{L}(G1,G2,D1,D2,D3,D4) = \mathcal{L}_{GAN}(G1(X),D1(Y))
+ \mathcal{L}_{GAN}(G2(Y),D2(X)) + \lambda_1 \mathcal{L}_{idt}(G1,G2)+\\
 \mathcal{L}_{cGAN}(G2(G1(X)),D3(G1(X),X)) + \mathcal{L}_{cGAN}(G1(G2(Y)),D4(G2(Y),Y))  + \lambda_{2}\mathcal{L}_{cyc \ell_1}(G1,G2)\nonumber   
\end{align}
\normalsize
$\lambda_1$ and $\lambda_2$ denote the corresponding weights for $\mathcal{L}_{idt}$ and $\mathcal{L}_{cyc \ell_1}$. Moreover, for paired data case we have experimented with perceptual loss using VGG19 pretrained network \cite{c34} as used in \cite{c19}.  

The major difference to previous literature (e.g. \cite{c3} and \cite{c1}) is the proposed four discriminator construction that work together with two generators to facilitate both paired and unpaired data simultaneously. The added structure enables new use cases, but also improves the performance in the case of only paired or unpaired data. We experimentally validate our model in the subsequent sections.  

\subsection{Network Architectures}
The architecture of \(G1\) and \(G2\) are same and contain 9-residual blocks \cite{c30}, pair of convolution block with stride -2 and another pair of fractionally strided convolution block with stride \(1/2\). The architecture of \(D1,\ D2,\ D3\) and \(D4\) are similar to PatchGAN \cite{c1}, \cite{c3}. However, \(D3\) and \(D4\) have provision to take the extra conditional information as their input channels. In our case for all the four discriminators, the size of the patch is \(70   \times70\). More details are given in the supplementary material.

\vspace{-0.3cm}
\subsection{Training Details}

Initially, all the datasets are resized to \(256\times256\times3\). During training, they are upscaled to \(286\times286\times3\) and random crops of size \(256\times256\times3\) are used as input to the model. First the paired data are applied to our model for \(50\) epochs and then unpaired data are applied for the rest from total epochs of 200. For the calculation of $\mathcal{L}_{cGAN}$ and $\mathcal{L}_{GAN}$, we have used least square loss \cite{c31} instead of log likelihood loss. Moreover, the discriminators are updated from a pool of 50 generated images instead of the most recent generated one. Similar strategies are used in \cite{c1},\cite{c32} to stabilize the training of GANs. For all the experiments, $\lambda_1$ and $\lambda_2$ in the loss function are 10 and 5 respectively. 
Our model is optimized with an Adam optimizer \cite{c33} and batch size is kept constant with 1. The starting learning rate for \(G1,G2\)  and \(D1,D2\) is kept at 0.0002 whereas for \(D3, D4\) is kept at 0.0001. For \(D3, D4\), the learning rate is fixed for all the epochs whereas the learning rate of others are kept at 0.0002 for 100 epochs and linearly decayed to zero in the remaining epochs.         

\section{Experiments}
In this section, we validate our method in several image-to-image translation setups using three different datasets: Cityscapes \cite{c4}, maps vs aerial images \cite{c1,c3}, and Mapillary Vistas \cite{c27}. The obtained results are qualitatively and quantitatively compared against two recent baseline methods: cycleGAN \cite{c1} and pix2pix \cite{c3} which utilise unpaired and paired training data respectively. These baselines were selected since they are the most relevant previous works with respect to our architecture and since they were shown to outperform many previous state-of-the-art methods in the corresponding publications. In the following sections we will explain the exact experiments and results for each dataset separately.

\begin{figure}[t]
\centering
\includegraphics[width=12.3cm,height=5.5cm]{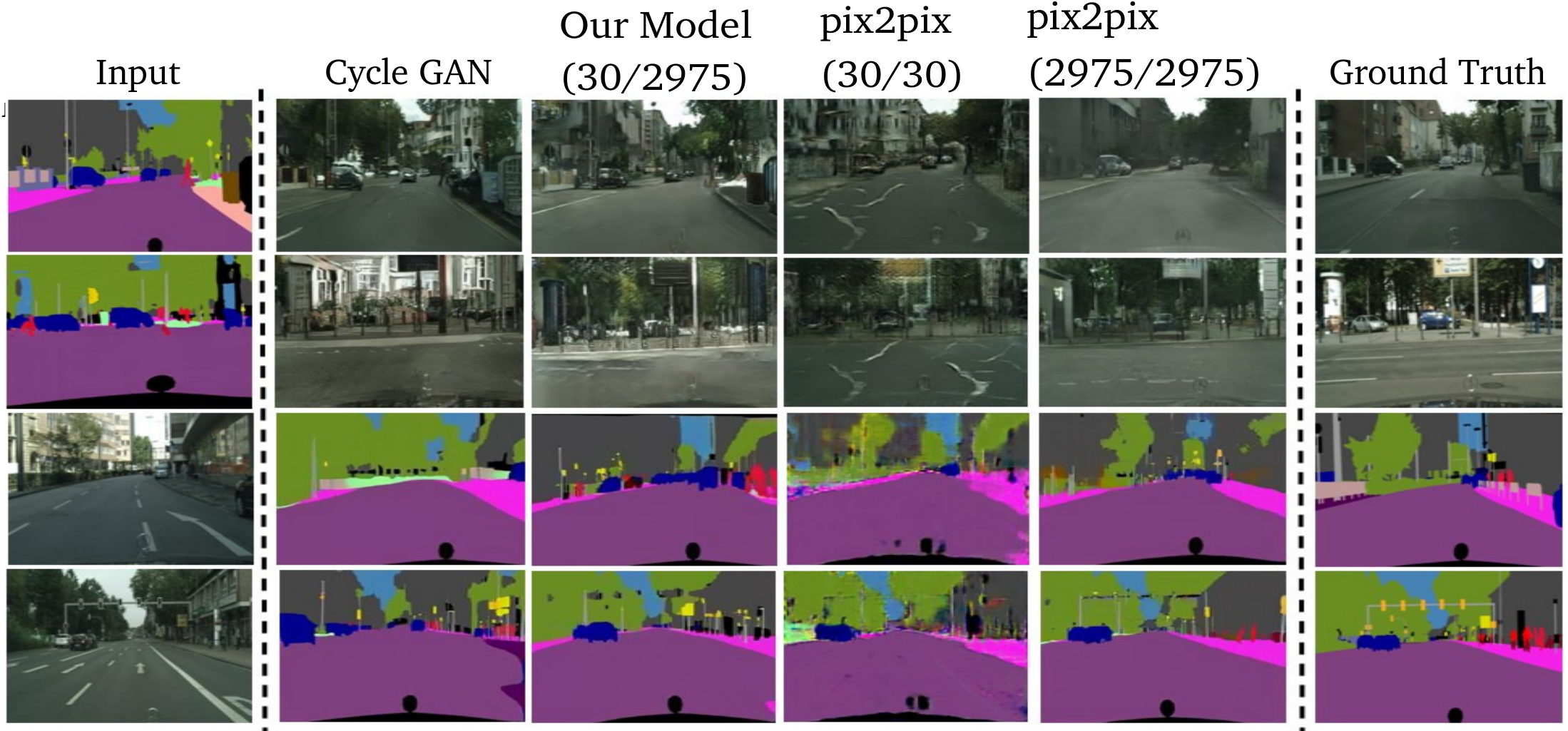}
\caption{Qualitative results for label map to photograph (first and second row) and photograph to label map (third and fourth row) translation tasks. The method corresponding to the results in each column is indicated above. The information in parenthesis denote the number of paired training examples and the total number of training data. We recommend the reader to zoom-in for observing the differences better.}
\label{fig5}
\end{figure}   

\vspace{-0.3cm}
\subsection{Translation between semantic label maps and RGB images}

Cityscapes \cite{c4} dataset consists of 3475 road scene images recorded in several locations in Germany. For urban scene understanding, these images are provided with the corresponding pixel level semantic annotations as shown in Figure \ref{fig1} and Figure \ref{fig5} as input and ground truth pair. The dataset is further split into training and validation parts containing 2975 and 500 samples respectively. 
We utilise Cityscapes to perform two types of translation tasks: 1) from semantic maps to RGB photographs and 2) from RGB photographs to semantic labels. Furthermore, we will modify the number of training samples for which the alignment between RGB images and label maps are utilised. This is done in order to assess how the number of paired vs.~unpaired samples affect the results. We are particularly interested in cases where there are relatively few paired samples (e.g. 50) and a large number of unpaired data.

\paragraph{Qualitative comparison}
Figure \ref{fig5} shows the qualitative results of our model and the baseline methods \cite{c1,c3}. The information in parenthesis denote the number of paired training samples and the total number of training samples. Note that since pix2pix \cite{c3} is not able to use unpaired training data, the corresponding total number of training samples is always equal to the number of paired samples. Similarly CycleGAN \cite{c1} uses only unpaired data. 

One may notice from the results that CycleGAN gets confused between tree and building labels. We observed that this kind of behaviour is very common throughout the dataset. Unfortunately, CycleGAN does not provide any easy way to include additional supervision to prevent such phenomenon. In contrast, our method can utilise supervision from aligned image pairs and we observed that already a small set of such pairs (e.g.~30) prevents this type of label switching problems. 

When comparing to pix2pix results, one can observe a clear benefit from the additional unpaired samples used in our method. Moreover, the visual quality of our results using pairing information for only 30 samples seems to be comparable to those obtained by pix2pix using pairing information for all 2975 training examples. One can find further example results from the supplementary material.  
 
\paragraph{Quantitative comparison}
In addition to the qualitative results, we also assess the methods quantitatively. To this end, we will follow the approach presented in \cite{c28}. For the first transformation task (labels to RGB), we apply a pretrained segmentation network \cite{c29} to the generated RGB images and compare the obtained label maps against the ground truth semantic segmentation maps. Following \cite{c28}, we compute pixel accuracy, mean accuracy and mean Intersection over Union (IU). In the case of RGB to label map translation, we compute the same metrics directly between our outputs and the corresponding ground truth maps. Note that the generated output label maps must be quantised before calculating the results. Table \ref{table1} contains the obtained results for different methods and training setups. 

In particular, our method seems to outperform the CycleGAN even if there are no paired training samples available. The key to this improvement is the proposed new adversarial cycle consistency loss. Furthermore, there is a clear additional improvement in the results when even a small set of paired training samples are added. Finally, we note that our method obtains similar results as pix2pix when pairing is provided for all training samples. We can observe the improvement over pix2pix when including a perceptual loss (VGG loss \cite{c19}) in our model.

\setlength{\tabcolsep}{1pt}
\begin{table}
\scriptsize 
\begin{center}
\begin{tabular}{l@{\hskip 0.17cm}l@{\hskip 0.2cm}l@{\hskip 0.1cm}l@{\hskip -0.2cm}l@{\hskip 0.3cm}l@{\hskip 0.1cm}l@{\hskip -0.2cm}l}
\hline
 &  &  & label$\rightarrow$photo &  &  & photo$\rightarrow$label & \\
Models & $\frac{paired}{total}$ & Pixel Acc. &Mean Acc. &Mean IU & Pixel Acc. & Mean Acc. &  Mean IU \\
\hline
Cycle GAN \cite{c1}  & -/2975 & 0.47 & 0.12 & 0.08 & 0.47 & 0.12 & 0.09 \\
Our Model & -/2975 & \textbf{0.56} & \textbf{0.16} & \textbf{0.12}  & \textbf{0.53} & \textbf{0.17} & \textbf{0.12}\\ 
\hline
pix2pix \cite{c3}  & 30/30 & 0.58 & 0.18 & 0.12  & 0.65 & 0.20 & 0.15\\
Our Model & 30/2975 & \textbf{0.66} & \textbf{0.19} & \textbf{0.14}  & \textbf{0.66} & 0.20 & 0.15 \\
\hline
pix2pix \cite{c3}  & 50/50 & 0.60 & 0.17 & 0.12  & \textbf{0.68} & \textbf{0.22} & \textbf{0.16}\\
Our Model & 50/2975 & \textbf{0.65} & \textbf{0.19} & \textbf{0.13}  & 0.67 & 0.20 & 0.15 \\
\hline
pix2pix \cite{c3}  & 70/70 & 0.62 & 0.18 & 0.13  & \textbf{0.70} & 0.21 & \textbf{0.17}\\
Our Model & 70/2975 & \textbf{0.67} & \textbf{0.19} & \textbf{0.14}  & 0.68 & 0.21 & 0.16 \\
\hline
pix2pix \cite{c3} & 2975/2975 & 0.68 & 0.21 & 0.15 & 0.75 & 0.28 & 0.21 \\
Our Model & 2975/2975 & 0.68 & 0.21 & 0.15  & 0.74 & 0.29 & 0.22\\
Our Model (VGG loss) & 2975/2975 & \textbf{0.70} & 0.21 & \textbf{0.16}  & \textbf{0.76} & \textbf{0.32} & \textbf{0.24}\\
\hline
\end{tabular}
\end{center}
\caption{Quantitative results for semantic label to photograph and photograph to semantic label translation tasks obtained using Cityscapes \cite{c4} dataset.}
\label{table1}
\vspace{-0.1cm}
\end{table}
\normalsize
\begin{figure}[t]
\centering
\includegraphics[width=12cm,height=4.1cm]{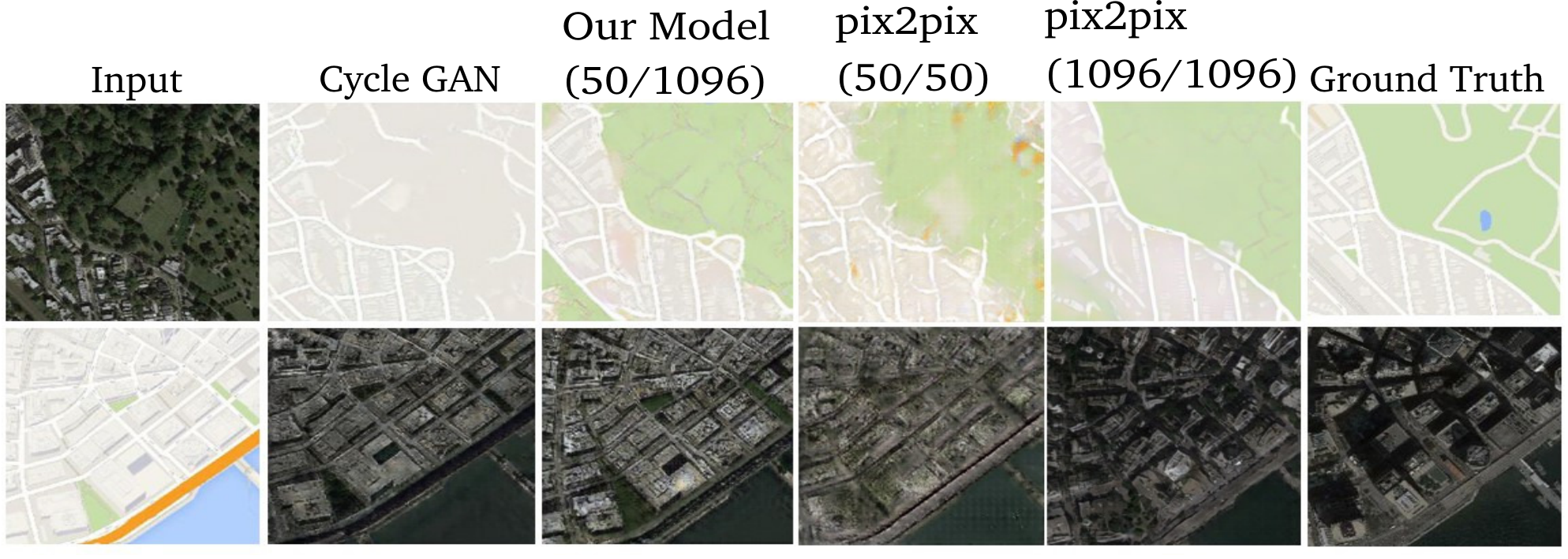}
\caption{Translation from aerial images to maps (first row) and vice versa (second row).}
\label{fig6}
\end{figure} 

\vspace{-0.5cm}
\subsection{Translation between aerial photographs and maps}
Aerial photograph and maps dataset proposed in \cite{c1,c3} consists of 1096 image pairs of aerial photographs and maps. As in the previous section, we use this dataset to learn transformations from photographs to maps and vice versa. Figure \ref{fig6} shows a few illustrative example results for our method and the baselines. One can refer to supplementary material for further examples. In this experiment, we show our results obtained using only 50 paired data out of 1096 training samples. However, other similar training setups lead to comparable results.
 
The obtained results indicate similar improvements over the baselines as in the case of label map to RGB image transformation. Our model successfully learns mappings for areas such as vegetation which are completely ignored by cycle GAN \cite{c1} (see Fig.\ref{fig6}). The other baseline, pix2pix, is not able to obtain comparable performance when only a small set of paired samples are available.  

\begin{figure}[t]
\centering

\includegraphics[width=12.2cm,height=6.3cm]{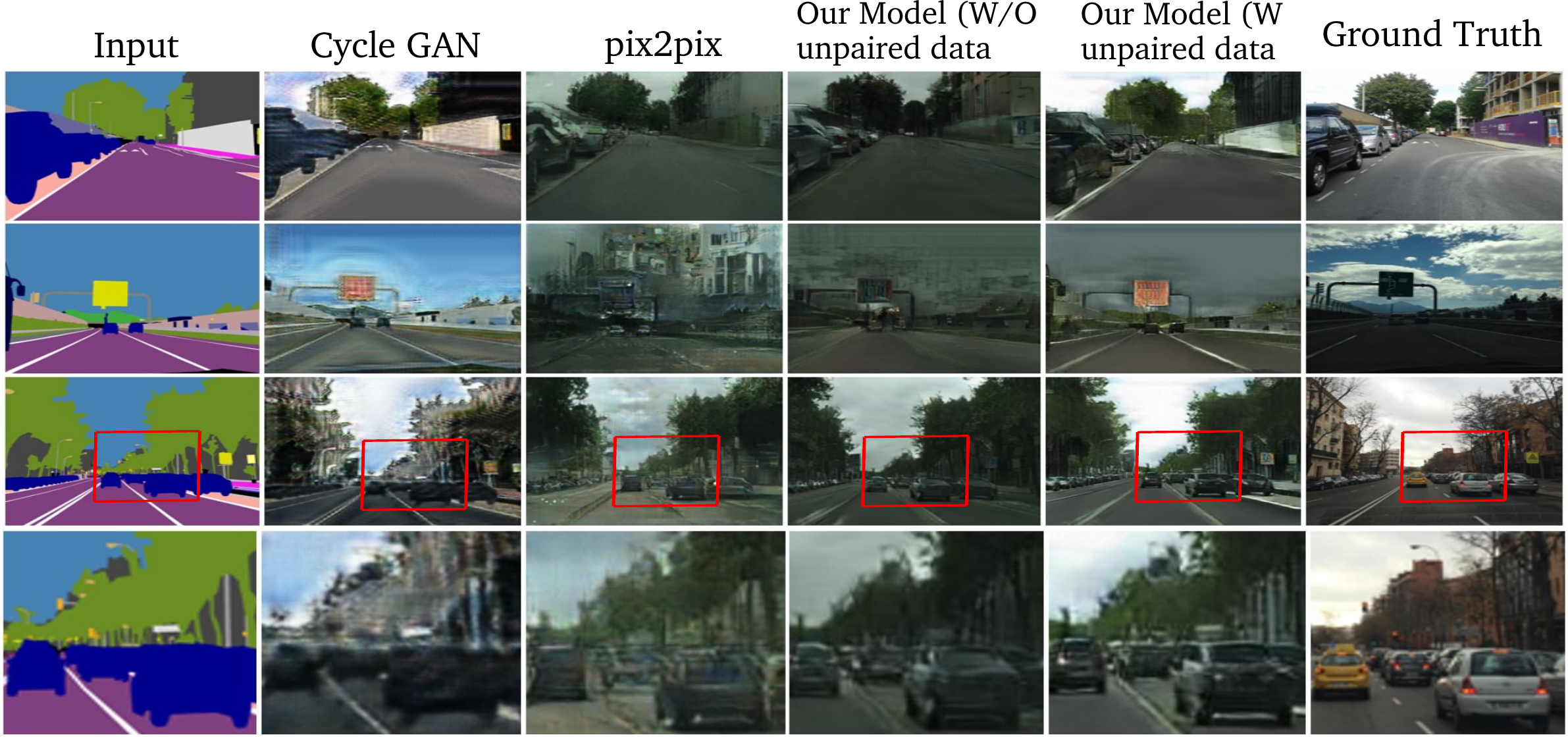}
\caption{Translation results from semantic label maps to RGB images on Mapillary Vistas dataset \cite{c27}. The method corresponding to the results in each column is indicated above. The last row shows a zoom-up corresponding to the red rectangle on the row above.}
\label{fig7}
\end{figure}


\subsection{Translation using supervision across datasets}

In addition to the previous experiments where train and test images originate from a single dataset, we assess the ability of our model to transfer knowledge from one dataset to another. For this purpose, we take all 2975 aligned training image pairs from the Cityscapes dataset \cite{c4} and an additional 100 unaligned example images\footnote{Note that it is not necessary to use the label maps from the Mapillary Vistas since the Cityscapes label maps can be used with Mapillary vistas RGB images as (unpaired) training examples.} from the Mapillary Vistas dataset presented in \cite{c27}. Mapillary Vistas contain similar road scene images and semantic label maps as Cityscapes. However, the Mapillary Vistas images are recorded from a much wider set of geographical locations around the world and therefore the images have substantially greater variability compared to Cityscapes. For instance, the road scenes in an African city may look quite different from those in Germany. 

Our task is to learn a translation model from semantic label maps to Mapillary Vistas style road scene images. Since we do not use any pairing information from Mapillary Vistas, the pix2pix model can be trained using only the Cityscape examples. On the other hand, CycleGAN can be trained using examples from both dataset. However, the results were clearly better if the model was trained using only Mapillary data as presented in Figure \ref{fig7}. 

For our method, we provide the results in two setups: 1) using only paired training samples from Cityscapes (same as pix2pix) and 2) using paired samples from Cityscapes and unpaired samples from Mapillary Vistas. 
The example results are provided in Figure \ref{fig7}. Our model seems to be better in terms of producing realistic textures and overall structure compared to the baselines. Furthermore, we observe a clear improvement from the additional unpaired Mapillary Vistas data indicating that it is possible leverage the supervision from similar dataset to improve the image-to-image translation results. One can find more example results in the supplementary material. 

\section{Conclusions}
We proposed a new general purpose method for learning image-to-image translation models from both aligned and unaligned training samples. Our architecture consists of two generators and four discriminators out of which two operate in dual role depending on type of the training instance. The advantage of this construction is that our model can utilise all training data irrespective of its type. We assessed the proposed method against two strong baselines presented recently. The results indicated both qualitative and quantitative improvements, even in the case where training data solely consists of paired or unpaired samples. To our knowledge, this is the first work to consider such hybrid setup in image-to-image translation.

\bibliographystyle{splncs}
\bibliography{egbib}
\newpage
\section*{Supplementary material}

In this material we provide details about the architecture and more qualitative results of our proposed model on cityscapes dataset \cite{c4}, aerial photos $\leftrightarrow$ maps dataset \cite{c3} and cross data evaluation on Mapillary vistas dataset \cite{c27} as discussed in section 4.

\begin{figure}
\centering
\includegraphics[width=12.3cm,height=13cm]{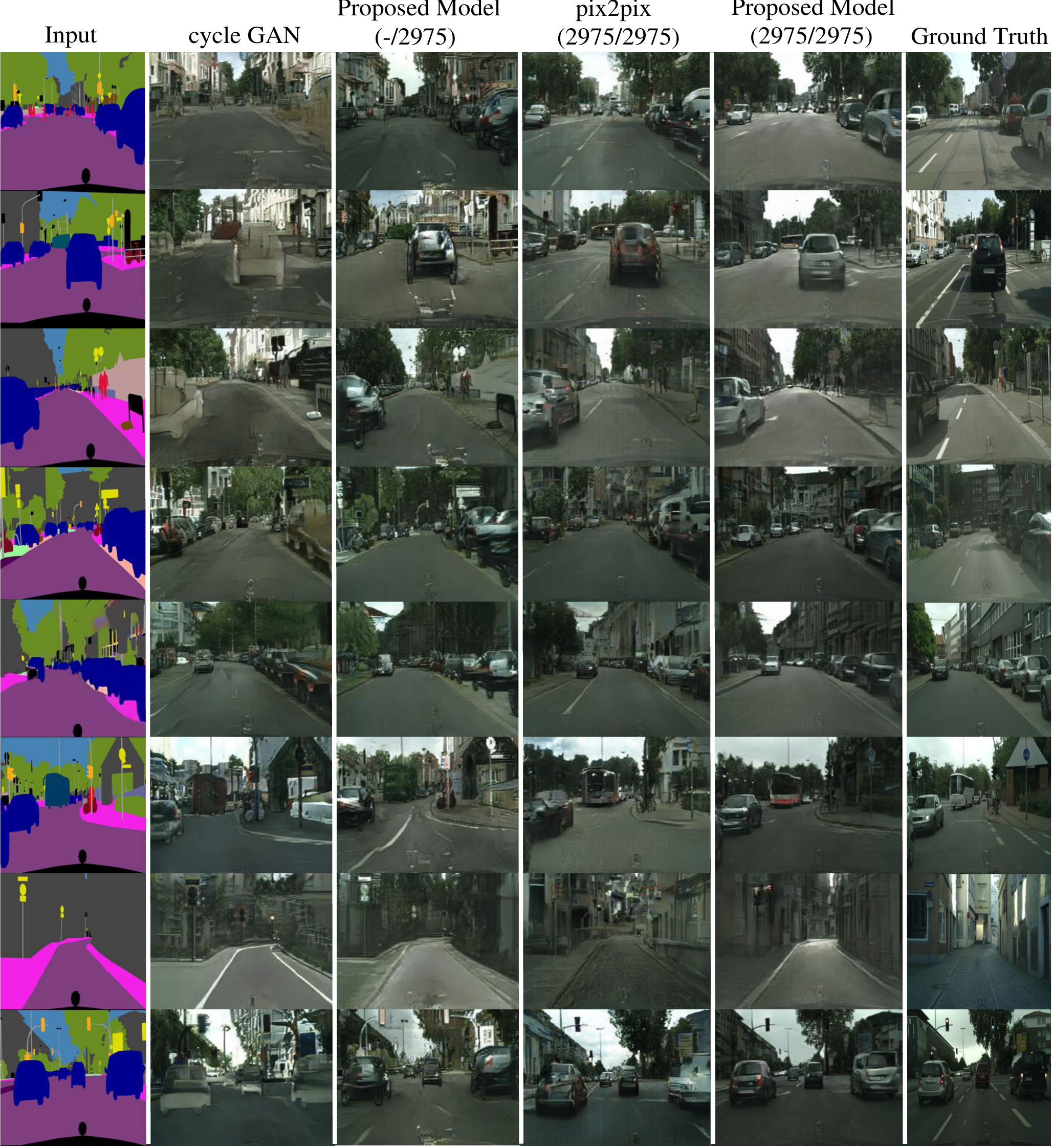}
\caption{Results obtained from various models for semantic labels to photo translation. From left to right: input, cycle GAN \cite{c1} (only unpaired training
examples), proposed model (only unpaired training
examples), pix2pix \cite{c3} (2975 paired training
examples), proposed model (2975 paired training
examples) and ground truth image.}
\label{sfig1}
\end{figure}

\begin{figure}
\centering
\vspace{-2.5cm}
\includegraphics[width=12.3cm,height=13.5cm]{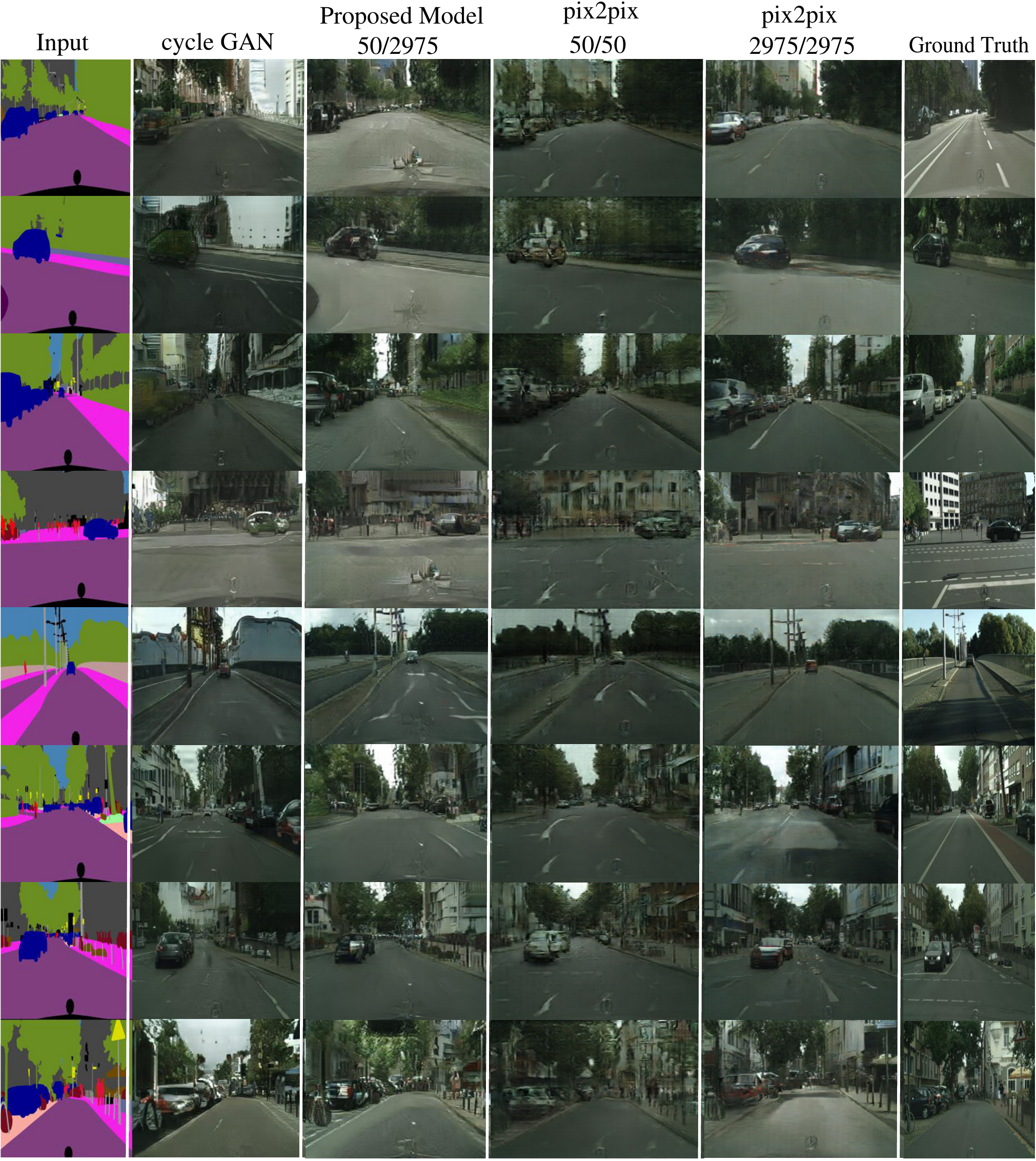}
\caption{Results obtained from various models for semantic labels to photo translation. From left to right: input, cycle GAN \cite{c1} (only unpaired training
examples), proposed model (50 paired data out of 2975 training samples), pix2pix \cite{c3} (only 50 paired training samples), pix2pix (2975 paired training samples) and ground truth image.}
\label{sfig2}
\end{figure}

\begin{figure}
\centering
\vspace{-2.5cm}
\includegraphics[width=12.3cm,height=13.5cm]{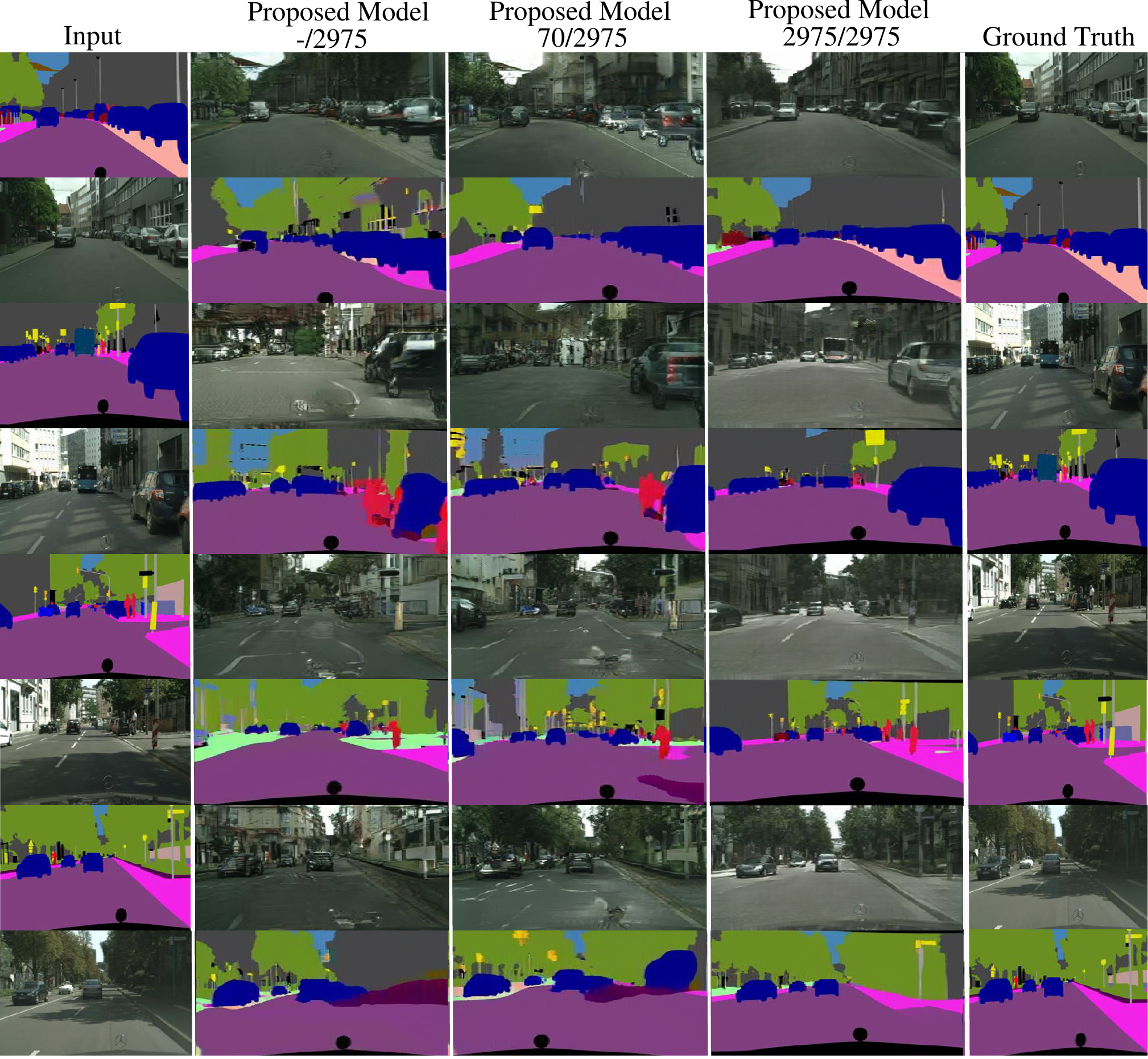}
\caption{Results obtained from proposed model for semantic labels $\leftrightarrow$ photo translation task. From left to right: input, proposed model (only unpaired training
examples), proposed model (70 paired data out of 2975 training samples), proposed model (2975 paired training
examples) and ground truth image.}
\label{sfig3}
\end{figure}

\paragraph{Translation between semantic label maps and RGB images}
Out of all the training images in the cityscapes dataset, various amount of pairing information (0, 50, 70, 2975 image pairs out of 2975 training images) between the domains are used and results are compared with baseline methods in Figure \ref{sfig1}, \ref{sfig2} and \ref{sfig3} for label $\leftrightarrow$ photo translation task.

\begin{figure}
\centering
\vspace{-2.5cm}
\includegraphics[width=12.3cm,height=13.5cm]{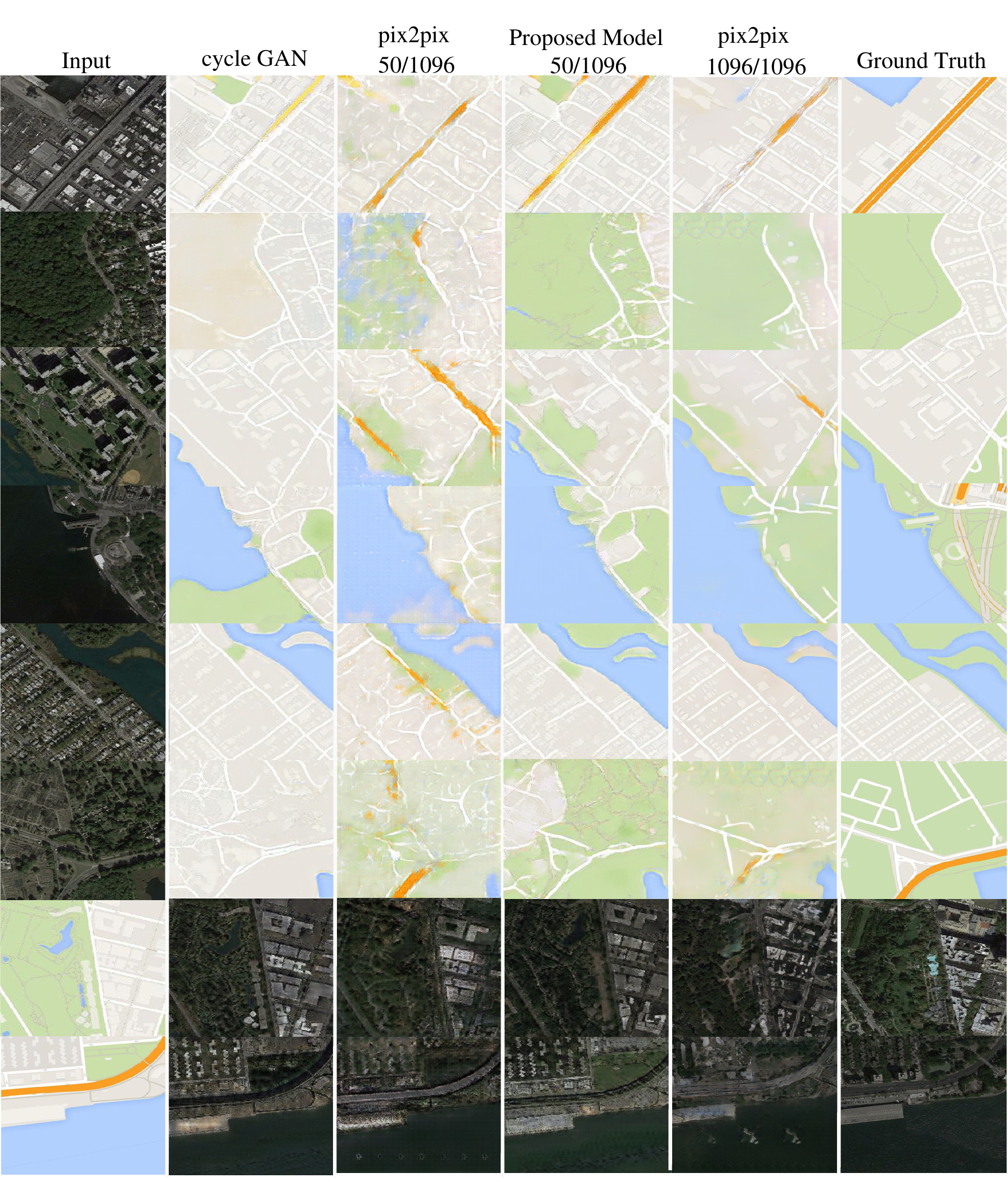}
\caption{Results obtained from various models for aerial photos $\leftrightarrow$ maps translation. From left to right: input, cycle GAN \cite{c1} (only unpaired training
examples), pix2pix \cite{c3} (only 50 paired training samples), proposed model (50 paired data out of 1096 training samples), pix2pix (1096 paired training samples) and ground truth image. }
\label{sfig4}
\end{figure}

\paragraph{Translation between aerial photographs and maps} 
In Figure \ref{sfig4}, we have provided similar image translation results for aerial photos $\leftrightarrow$ maps by using only 50 paired data out of total 1096 training images.

\begin{figure}
\centering
\vspace{-2.5cm}
\includegraphics[width=12.3cm,height=14cm]{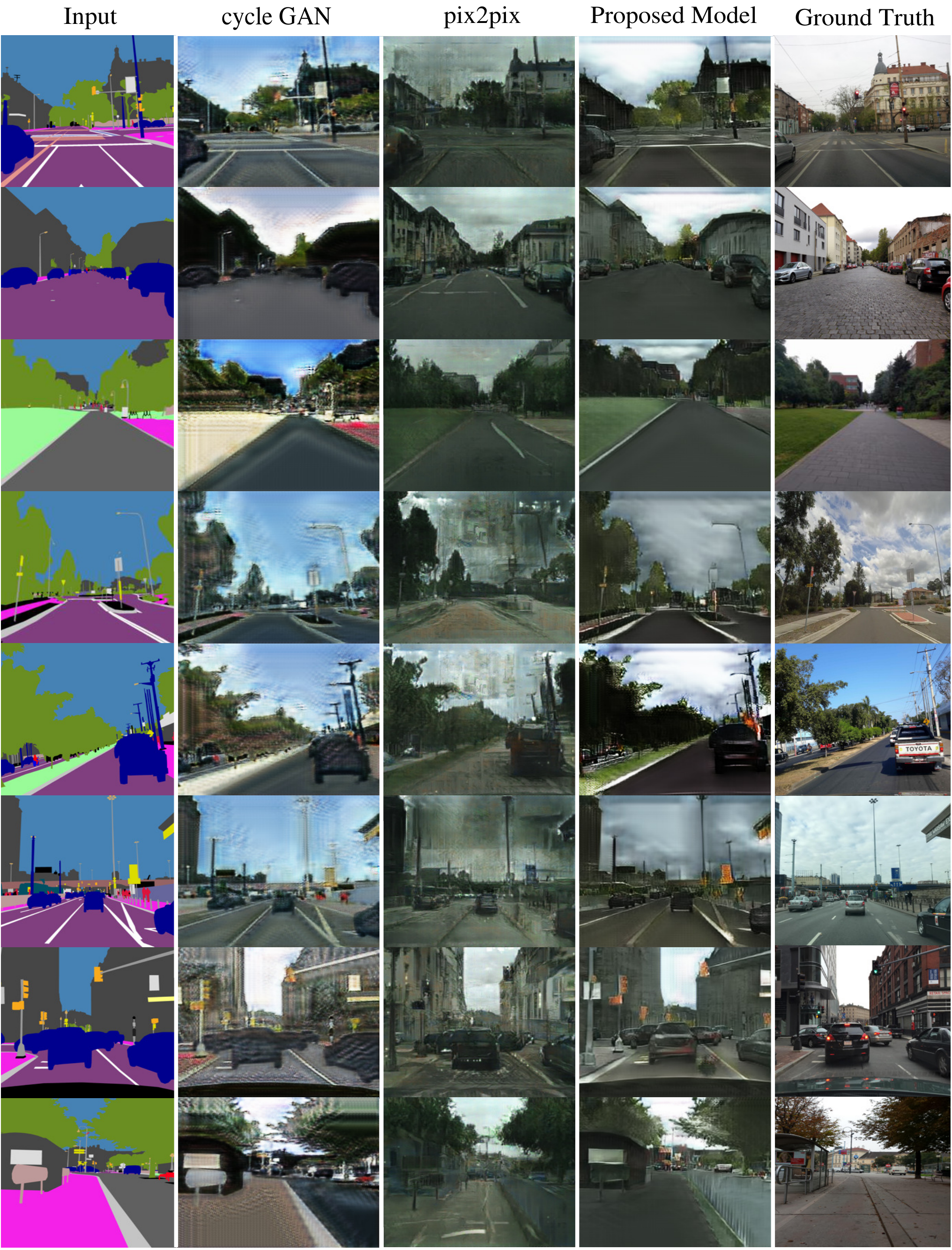}
\caption{Results obtained from various models on Mapillary vistas dataset for label $\rightarrow$ photo translation. From left to right: input, cycle GAN \cite{c1}, pix2pix \cite{c3}, proposed model and ground truth image.}
\label{sfig5}
\end{figure}

\paragraph{Translation using supervision across datasets}
For the cross data supervision, our training set consists of 2975 paired images from cityscapes \cite{c4} and 100 unpaired images from Mapillary vistas dataset \cite{c27}. All the models are tested on Mapillary vistas images that are not involved in training. For the pix2pix \cite{c3} or cycle GAN \cite{c1}, only paired or unpaired data are used for the training respectively and results are shown in Figure \ref{sfig5}. First our model is trained on paired data from cityscapes and then fine tuned on Mapillary vistas unpaired data to produce high quality results than the baseline methods as shown in Figure \ref{sfig5}.

\paragraph{Architecture}

We adopt the architecture and naming convention form \cite{c1} for 256 $\times$ 256 resolution images and used the training scheme from the Zhu et.al's Github repository \footnote{https://github.com/junyanz/pytorch-CycleGAN-and-pix2pix}.

Let, $c7s1-k$: convolution-instancenorm-ReLU layer with $k$ filters of length 7 $\times$ 7 and stride 1. $dk$: convolution-instancenorm-ReLU layer with $k$ filters of length 3 $\times$ 3 and stride 2. $Rk$:
residual block with equal number of filters in two convolutional layers of filter size 3 $\times$ 3. $uk$: fractional-strided-Convolution-
InstanceNorm-ReLU layer with $k$ filters of length 3 $\times$ 3 and stride 1/2. $Ck$: convolution-instancenorm-LeakyReLU layer with $k$ filters of length 4 $\times$ 4 and stride 2. Slope of LeakyRelU is kept at 0.2 and as a padding scheme reflection padding is used. \\
\\
\textbf{ Generator Architecture}\\
$c7s1-64,\ d128, \ d256, \ R256, \ R256, \ R256,
\ R256, \ R256, \ R256, \ R256, \ R256, \ R256, \\ \ u128, \ u64,
\ c7s1-3$ \\
\\
\textbf{Discriminator Architecture}\\
$C64-C128-C256-C512$ \ 

For only $C64$, instance norm is not applied and after the last layer, we converted the output to one dimensional by applying a convolution layer.  

\end{document}